\documentclass{article} 
\usepackage{colm2024_conference}

\usepackage[utf8]{inputenc} 
\usepackage[T1]{fontenc}    
\usepackage{hyperref}       
\usepackage{url}            
\usepackage{booktabs}       
\usepackage{amsfonts}       
\usepackage{nicefrac}       
\usepackage{microtype}      
\usepackage{xcolor}         
\usepackage{amsmath}
\usepackage{graphicx}
\usepackage{subfigure}
\usepackage{natbib}
\usepackage{amsthm}
\usepackage{amssymb}
\usepackage{algorithm}
\usepackage{algpseudocode}
\definecolor{royalblue}{rgb}{0.25, 0.41, 0.88}
\usepackage{enumitem}

\newtheorem{theorem}{Theorem}[section]

\theoremstyle{definition}
\newtheorem{definition}{Definition}[section]

\title{Automatically Identifying Local and Global Circuits with Linear Computation Graphs}

\colmfinalcopy

\author{
Xuyang Ge$^1$ \hspace{.3em}
Fukang Zhu$^1$ \hspace{.3em}
Wentao Shu$^1$ \hspace{.3em}
Junxuan Wang$^1$ \hspace{.3em}
Zhengfu He$^1$\thanks{This work lies in the OpenMOSS Mech Interp project led by Zhengfu He.}\hspace{.4em}
Xipeng Qiu$^1$\thanks{Corresponding author.} \hspace{.3em}
\\
[1ex]
\texttt{xyge20@fudan.edu.cn \hspace{.3em} zfhe19@fudan.edu.cn}
\\
[1ex]
$^{1}$Open-MOSS Team, Fudan Unversity \\
}

\fancypagestyle{firstpage}{
  \lhead{\begin{picture}(0,0)\put(0,-6){\includegraphics[width=0.3\linewidth]{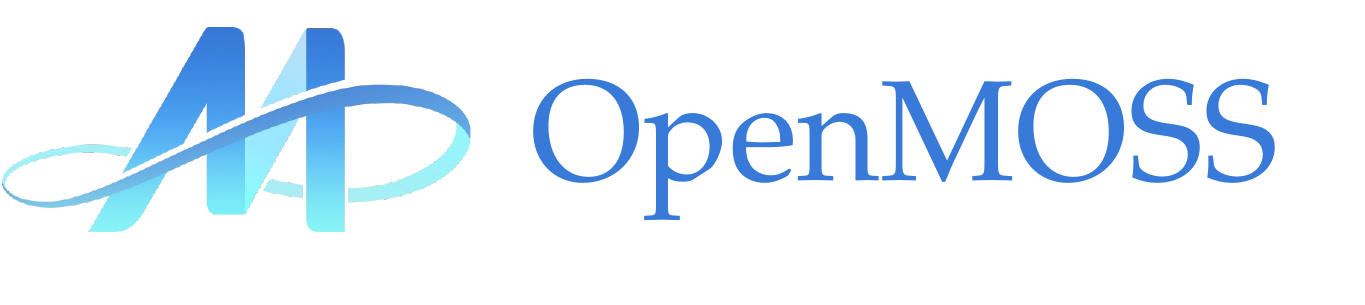}}\end{picture}}
}

\begin{document}

\maketitle

\thispagestyle{firstpage}

\begin{abstract}
  Circuit analysis of any certain model behavior is a central task in mechanistic interpretability. We introduce our circuit discovery pipeline with Sparse Autoencoders (SAEs) and a variant called Transcoders. With these two modules inserted into the model, the model's computation graph with respect to OV and MLP circuits becomes strictly linear. Our methods do not require linear approximation to compute the causal effect of each node. This fine-grained graph identifies both end-to-end and local circuits accounting for either logits or intermediate features. We can scalably apply this pipeline with a technique called Hierarchical Attribution. We analyze three kinds of circuits in GPT-2 Small: bracket, induction, and Indirect Object Identification circuits. Our results reveal new findings underlying existing discoveries.
\end{abstract}

\section{Introduction}
Recent years have seen the rapid progress of mechanistically reverse engineering Transformer language models~\citep{Vaswani2017Transformer}. Conventionally, researchers seek to find out how neural networks organize information in its hidden activation space~\citep{Olah2020early-vision, Gurnee2023haystack, Zou2023REPE} (i.e. features) and how learnable weight matrices connect and (de)activate them~\citep{Olsson2022induction, Wang2023IOI, Conmy2023ACDC} (i.e. circuits). One fundamental problem of studying attention heads and MLP neurons as interpretability primitives is their polysemanticity, which under the assumption of linear representation hypothesis is mostly due to superposition~\citep{Elhage2022superposition, Larson2023ExpandingSuperposition, Greenspan2023AttentionSuperposition}. Thus, there is no guarantee of explaining how these components impact model behavior out of the interested distribution. Additionally, circuit analysis based on attention heads is coarse-grained because it lacks effective methods to explain the intermediate activations. 

Probing~\citep{Alain2017Probe} in the activation for a more fine-grained and monosemantic unit has succeeded in discovering directions indicating a wide range of abstract concepts like truthfulness~\citep{Li2023ITI} and refusal of AI assistants~\citep{Zou2023REPE, Arditi2024refusal}. However, this supervised setting may not capture features we did not expect to present.

Sparse Autoencoders (SAEs)~\citep{Bricken2023monosemanticity, Cunningham2023SAE} have shown their potential in extracting features from superposition in an unsupervised manner. This opens up a new perspective of understanding model internals by interpreting the activation of SAE features. It also poses a natural research question: \textbf{how to gracefully leverage SAEs for circuit analysis?} Compared to prior work along this line~\citep{Cunningham2023SAE, He2024OthelloCircuit, Marks2024SparseFeatureCircuit}, our main contributions are as follows.

\begin{figure}
    \centering
    \includegraphics[width=.8\linewidth]{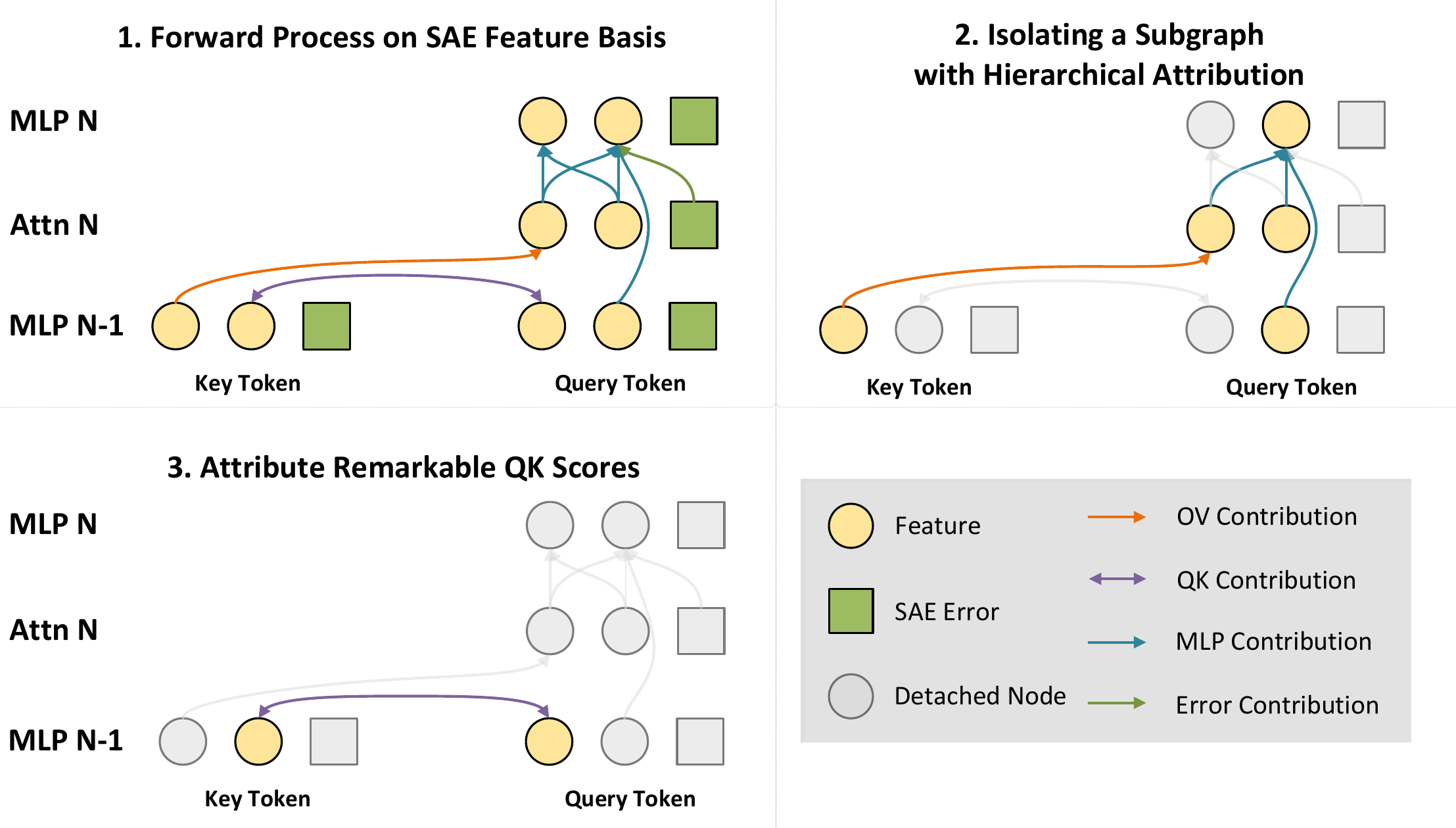}
    
    \caption{Overview of our method. For a given input, we (1) run forward pass once with MLP computation replaced by Trans. (2) Then a subgraph is isolated for a given input with \textit{Hierarchical Attribution} in one backward. (3) We then interpret important QK attention involved in the identified circuit.}
\end{figure}

\begin{itemize}
    \item We propose to utilize Transcoders, a variant of Sparse Autoencoders, to sparsely approximate the computation of MLP layers. This extends the linear analysis of Transformer circuits~\citep{elhage2021mathematical, He2024OthelloCircuit}.

    \item For a given input, OV + Transcoder (i.e., MLP) circuits strictly form a \textbf{Linear Computation Graph} without linear approximation of any non-linear function. This precious linearity enables circuit discovery and evaluation with only one forward and one backward. 
    
    \item We propose \textit{Hierarchical Attribution} to isolate a subgraph of the aforementioned linear graph in an automatic and scalable manner. 

    \item We present a specific example in our analysis that offers more detailed insight into how each single SAE feature contributes to a desired behavior, e.g., forms a crucial QK attention or linearly activates a subsequent node in the computation graph. Such observations are not reported by existing work studying circuits in coarser granularity.

\end{itemize}

\section{Linear Computation Graphs Connecting SAE Features}
\label{sec:methodology}

\subsection{Sparse Autoencoder Features as Analytic Primitives}
\label{sec:sparse_autoencoder}

Sparse Autoencoder (SAE) is a recently emerging method to take features of model activation out of superposition~\citep{Elhage2022superposition}. Existing work has suggested empirical success in the interpretability of SAE features concerning both human evaluation~\citep{Bricken2023monosemanticity} and automatic evaluation~\citep{Bills2023autointerp}.

Concretely, an SAE and its optimization objective can be formalized as follows:

\begin{equation}
    \label{eq:sae}
    \begin{aligned}
        f&=\operatorname{ReLU}(W_Ex+b_E)\\
        \hat{x}&=W_Df\\
        \mathcal{L}&=\lVert x-\hat{x}\rVert_2^2+\lambda\lVert f\rVert_1,
    \end{aligned}
\end{equation}

where $W_E\in\mathbb{R}^{d_\text{SAE}\times d_\text{model}}$ is the SAE encoder weight, $b_E\in\mathbb{R}^{d_\text{SAE}}$ encoder bias, $W_D\in\mathbb{R}^{d_\text{model}\times d_\text{SAE}}$ decoder weight, $x\in\mathbb{R}^{d_\text{model}}$ input activation. $\lambda$ is the coefficient of L1 loss for balance between sparsity and reconstruction. We refer readers to Appendix~\ref{appendix:sae_training} for implementation details.

We train Sparse Autoencoders on GPT-2~\citep{Radford2019GPT2} to decompose \textit{all modules that write into the residual stream} (i.e. Word Embedding, Attention output and MLP output). Then, we can derive how a residual stream activation is composed of SAE features:

\begin{equation}
    x=\sum_{\mathcal{S}\in\text{Upstream SAEs}}\left(\sum_{i=1}^{d_\text{SAE}}f_i^\mathcal{S}{W_D^\mathcal{S}}_i+\varepsilon^\mathcal{S}\right) + p,
\end{equation}

where $f_i^\mathcal{S}$ and $\varepsilon^\mathcal{S}$ are feature activation and SAE error term of each upstream SAE $\mathcal{S}$. $p$ is the positional embedding of the current token. Since all submodules read and write into the residual stream, such a partition is crucial to connect upstream SAE features to downstream ones.
 
\subsection{Tackling MLP Non-linearity with Transcoders}
\label{sec:skip_sae}

    

The denseness and non-linearity of MLP in Transformers make sparse attribution of MLP features difficult. Since MLP activation functions have a privileged basis~\citep{Elhage2023privileged}, computation of MLP non-linearity must go through such an orthogonal basis of the MLP hidden space. There is no guarantee of observing sparse and informative correspondence between MLP neurons and learned SAE features. This annoying non-linearity cuts off the connection of upstream SAE features and MLP output (with linear algebraic operations).

To tackle this problem, we develop a new method called Transcoders to get around the MLP non-linearity. Transcoders are generalized forms of SAEs, which decouple the input and output of SAEs and allow for predicting future activations given an earlier model activation. Transcoders take in the pre-MLP activation and yield a sparse decomposition of MLP output. Formally, a Transcoder and its optimization objective can be written as:

\begin{equation}
    \begin{aligned}
        f&=\operatorname{ReLU}(W_Ex+b_E)\\
        \hat{y}&=W_Df\\
        \mathcal{L}&=\lVert y-\hat{y}\rVert_2^2+\lambda\lVert f\rVert_1,
    \end{aligned}
\end{equation},

which only differs from those of an SAE (Eq.~\ref{eq:sae}) by the label activation $y\in\mathbb{R}^{d_\text{model}}$ unbound with input activation $x$.

\paragraph{Key difference between Transcoders and MLP} We may find Transcoders and MLP with similar architecture: both are two fully connected blocks interspersed with an activation function. It's natural to ask why the non-linear activation function in MLP is deemed as an obstacle in circuit analysis but that in Transcoders is allowed. The key difference is that by constraining the sparsity, Transcoders neurons (which are just features) have an \textit{interpretable basis}. When computing how upstream feature $f_i^\mathcal{S}$ contributes to activated downstream feature $f_j^\mathcal{T}$ of Transcoder $\mathcal{T}$, it holds that $f_j^\mathcal{T}=f_i^\mathcal{S} \left(W_E^\mathcal{T} W_D^\mathcal{S}\right)_{ji}$. The $\left(W_E^\mathcal{T} W_D^\mathcal{S}\right)_{ji}$ part remains constant across inputs, which leads to an \textbf{edge invariance} between upstream and downstream features. 

Intuitively, this means when a main upstream contributor to a downstream feature has been activated in a different input, we can largely expect this downstream feature to be activated again unless some new resistances (upstream features with negative edges) have also been introduced. 

In contrast, we cannot find such invariant edges through MLP. Any connection from upstream to MLP output is indefinite, so we could only find linear approximations to measure these connections under local changes.




\subsection{QK and OV Circuits Are Independent Linear Operators on SAE Features}
\label{sec:attn_contribution}

QK and OV circuits account for how tokens attend to one another and how information passes to downstream layers, respectively. The linearity and independence of these two components have been widely discussed in previous work~\citep{elhage2021mathematical, He2024OthelloCircuit}. Specifically, QK circuits serve as a bilinear operator of any two residual streams w.r.t token $i$ and $j$:

\begin{equation}
    \begin{aligned}
        \operatorname{AttnScore}^h(x)_{ij}
        &=x_i{W_Q^h}^TW_K^hx_j^T\\
        &=\sum_{\mathcal{S}, \mathcal{T}\in\text{Upstream SAEs}}\sum_{p=1}^{d_\text{SAE}}\sum_{q=1}^{d_\text{SAE}} f_{i,p}^\mathcal{S} {W_D^\mathcal{S}}_p {W_Q^h}^T W_K^h {W_D^\mathcal{T}}_q^T f_{j,q}^\mathcal{T},\\
    \end{aligned}
\end{equation}

where $f_{i,p}$ means the activation of the feature $p$ at token $i$, and $W_Q^h, W_K^h$ are a given head $h$'s the query and key transformation. This decomposition shows how every pair of upstream features contributes to the attention score, making tokens containing critical information get attended. 

Once the attention score is determined, we can then move on to the OV circuits, which apply a linear transformation to all past residual streams and take a weighted sum:

\begin{equation}
    \label{eq:attn}
    \begin{aligned}
        \operatorname{Attn}(x)_i&=\sum_h\operatorname{AttnOutput}^h(x)_i\\
        &=\sum_h\sum_j\textcolor{royalblue}{\operatorname{AttnPattern}^h(x)_{i,j}} W_O^h W_V^h x_j,
    \end{aligned}
\end{equation}

where $W_O^h, W_V^h$ are a given head $h$'s output and value transformation. With \verb|AttnPattern| determined in the QK circuits, how upstream features affect downstream are successively determined since $W_O^h W_V^h$ is invariant.

From an input-independent perspective, the quadratic coefficient ${W_D^\mathcal{S}}_p {W_Q^h}^T W_K^h {W_D^\mathcal{T}}_q$ shows how feature pairs co-work for every attention score. Then, ${W_E^\mathcal{S}}_p W_O^h W_V^h {W_D^\mathcal{T}}_q$ (obtained by adding SAE encoder and decoder terms to Eq.~\ref{eq:attn}) determines the edge connecting upstream features and attention output features under a specific attention pattern. This two-step paradigm gives us a simplified and feature-based version of attention functionality and allows a fine-grained analysis through attention in a non-approximated manner.

In real-world applications, we often want to attribute an interested output (e.g., logits) to filter out critical features, which is a backward procedure. For the sake of a linear and exact attribution result, we can reverse the above two-step paradigm and 1) attribute through OV + Transcoder circuits and then 2) select important attention, attribute its attention score 
through the current QK and once again the upstream OV + Transcoder circuits (showed in Figure.~\ref{fig:hierachical-attribution-workflow}). The second step may be repeated several times to attribute attention important to another attention.

\section{Isolating Interpretable Circuits with Hierarchical Attribution}
\label{sec:hierarchical_attribution}

\begin{figure}
    \centering
    \subfigure[Workflow of performing \textit{Hierarchical Attribution} and standard attribution.]{
        \label{fig:hierachical-attribution-workflow}
        \includegraphics[width=0.48\linewidth]{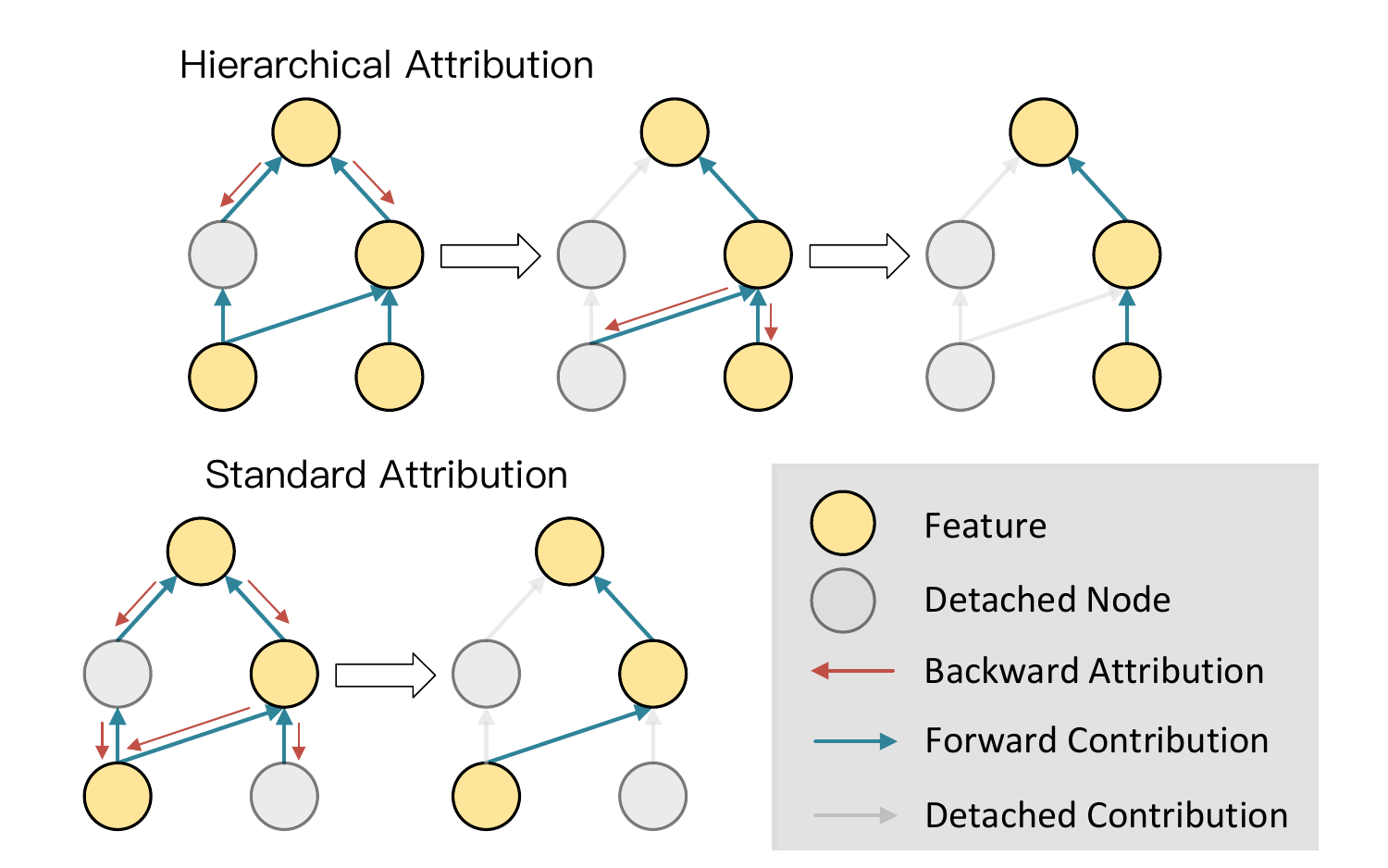}
    }
    \subfigure[Comparison between \textit{Hierarchical Attribution} and standard attribution.]{
        \label{fig:logit-recovery}
        \includegraphics[width=0.48\linewidth]{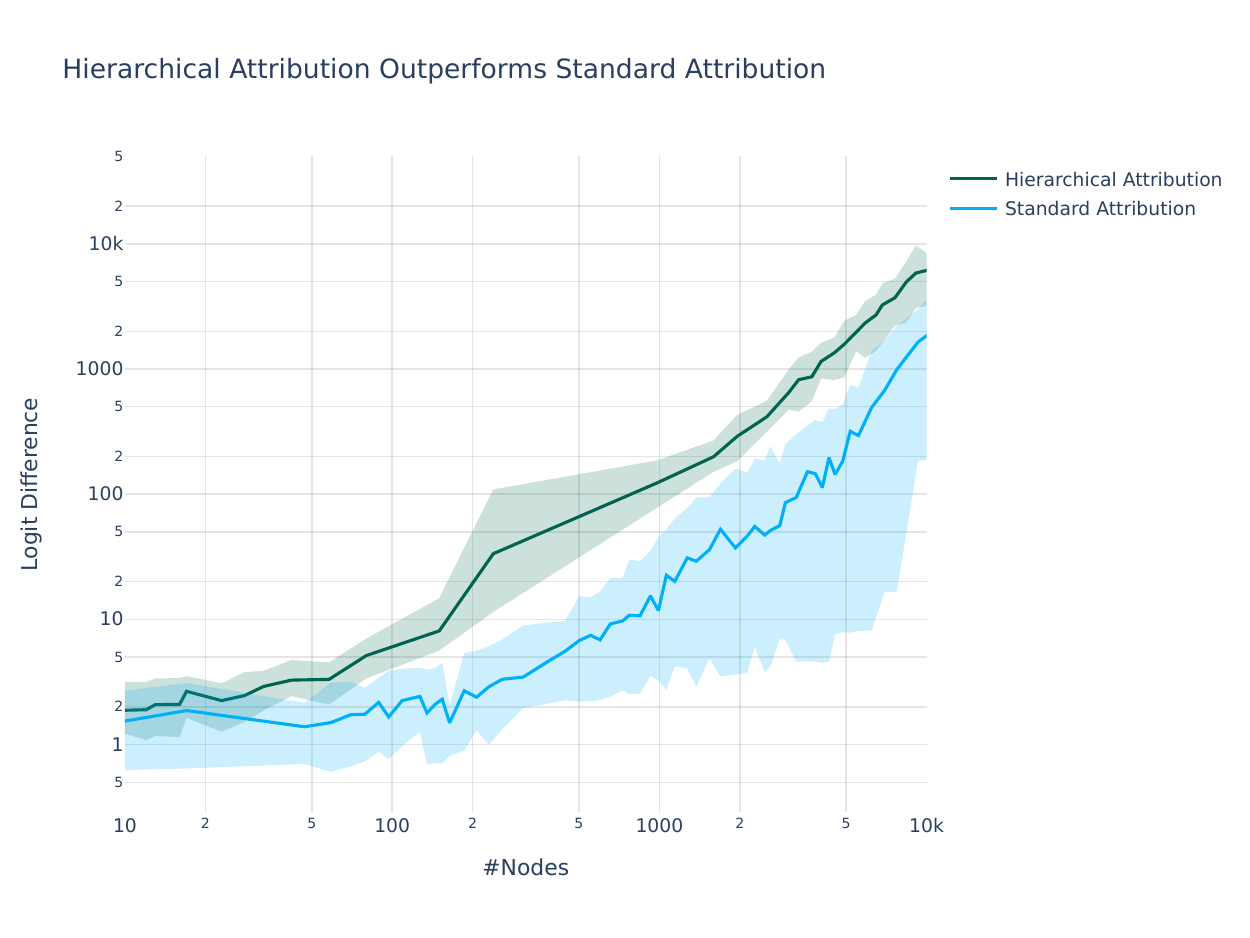}
    }
    
    \caption{Our \textit{Hierarchical Attribution} detaches unrelated nodes immediately after they receive gradient and stops their backpropagation, while standard attribution detaches nodes after the backward pass is completed. (Figure~\ref{fig:hierachical-attribution-workflow}). We sweep the number of remaining nodes, i.e., sparsity, and compare the logit recovery, i.e., faithfulness of the identified subgraph. Experiments are conducted on 20 IOI samples (See Section~\ref{sec:ioi_circuits}) across 30 sparsity thresholds. Results in Figure~\ref{fig:logit-recovery} show that \textit{Hierarchical Attribution} consistently outperforms standard attribution.}
\end{figure}

We have now obtained a linear computation graph including all OV and MLP modules, reflecting the model's internal information flow. This section introduces how to isolate and evaluate a subgraph of the key SAE features related to any interested output.

\paragraph{Formulation}

We are given a linear computation graph $G=(V, E)$, which is a directed acyclic graph. Each node $v\in V$ refers to an activated feature in the model forward pass. The node weight $a_v$ refers to the \textit{activation} of node $v$. Each edge $v\to u\in E$ represents that $a_v$ linearly affects $a_u$ by the edge weight $k_{v,u}$. For any non-leaf node $u$, its activation is completely determined by its direct predecessors, i.e., $a_u = \operatorname{ReLU}\left(\sum_{v\to u\in E} k_{v,u} a_v\right)$.

The term linear computation graph means every edge in the graph represents a linear function (under fixed attention scores). This guarantees a one-hop linear effect of activated features. It's not necessary that indirect effects between any two nodes are still linear since we allow a \verb|ReLU| gate inside the nodes, stopping unactivated nodes from forwarding further.


\paragraph{Two Types of Leaf Nodes}

We denote word embedding SAE features and the position embedding as \textit{interpretable leaf nodes}\footnote{We notice that not all SAE features are interpretable. We adopt a series of methods to improve the interpretability of SAEs further. See Appendix~\ref{appendix:sae_training}}. SAE errors also have zero in degree, but we cannot establish any explanation for these nodes. Thus, we call them \textit{uninterpretable leaf nodes}.

\paragraph{Isolating a Subgraph with Node Detaching} 
We prune unrelated nodes in the original linear computation graph to identify a subgraph accounting for the desired output.

\begin{definition}[Detaching a node]
The operation of detaching a node $v$ from graph $G$ is to get an induced subgraph $G'=G[V/v]$, which removes $v$ and all edges connecting to $v$ from $G$.
\end{definition}

We first need to detach all SAE errors since they cannot be interpreted, despite their empirically positive correlation to model performance~\citep{Gurnee2024pathological}. In the rest of the graph, with all leaf nodes being \textit{interpretable leaf nodes}, we need to detach nodes unrelated to the task.

\paragraph{Manual Pruning with Direct Contribution} For graphs with a small number of nodes, a simple solution is to manually inspect the interpretation of SAE features and their causal relation. This is often useful in understanding local behaviors but may be labor-intensive at scale.

\paragraph{Automatic Circuit Discovery with Hierarchical Attribution} 

We present how to perform scalable circuit discovery on this linear computation graph with gradient-based attribution~\citep{Kramar2024AtP}. 

\begin{definition}[Attribution Score]
The attribution score of node $v$ w.r.t. an interested output node $t$ is $\operatorname{attr}_{v,t}:=a_v\cdot\nabla_{a_t}a_v$.
\end{definition}


A natural idea would be running backward once and detaching nodes with $\operatorname{attr}_{v,t}$ lower than a given threshold $\tau$, as adopted in most prior work~\citep{Conmy2023ACDC, Marks2024SparseFeatureCircuit}. We propose to operate a breadth-first search style attribution pipeline we call \textit{Hierarchical Attribution}. 

\textit{Hierarchical Attribution} \textbf{detaches nodes on backward pass} instead of after backward, as shown in Figure~\ref{fig:hierachical-attribution-workflow} and a pseudo-code implementation in Appendix~\ref{appendix:hierachical_attribution}. When performing model backward, we stop the gradient propagation of any node $v$ that has $\operatorname{attr}_{v,t} < \tau$. This affects the attribution score of all predecessors of $v$. After we finish the backward propagation, all nodes with gradients make up our desired subgraph. Intuitively, attribution through detached nodes should not be taken into account; otherwise, their effect depends on excluded nodes in the final subgraph.

\paragraph{Evaluation}

We leverage a good property of linear graphs to evaluate identified circuits. 

\begin{theorem}
\label{theorem:attribution_equality}
For any subgraph $G'=G[V/v]$, the node weight of the root node is the sum of the attribution scores of all leaf nodes.
$$
    a_t = \sum_{\deg_{\text{in}}(v)=0} \operatorname{attr}_{v,t}
$$
\end{theorem}

We refer readers to Appendix~\ref{appendix:attribution_equality} for the proof.

This theorem allows us to instantly obtain how much $G'=G[V/v]$ accounts for the root node activation after we finish the pruning. Besides efficiency, another advantage of such evaluation is that it derives the causal effect of circuits without any intervention in the forward pass. It saves circuit evaluation from backup behaviors~\citep{Wang2023IOI} (also known as hydra effects~\citep{McGrath2023Hydra}) due to ablation.

In Figure~\ref{fig:logit-recovery}, we empirically validate the advantage of \textit{Hierarchical Attribution} over the standard attribution method in Indirect Object Identification circuit discovery~\citep{Wang2023IOI}.

\section{Attributing Intermediate SAE Features}
\label{sec:local_circuits}

An exciting application of Sparse Autoencoders is that they serve as \textit{unsupervised feature extractors} in the vast hidden activation space. This opens up opportunities for understanding intermediate activations and local circuit discovery, i.e., identifying a subgraph activating a given SAE feature instead of end-to-end circuits.

\subsection{How Transformers Implement In-Bracket Features}

\begin{figure}
    \centering
    \subfigure[Formation of In-Bracket Features]{
        \label{fig:in-bracket-information-flow}
        \includegraphics[width=0.6\linewidth]{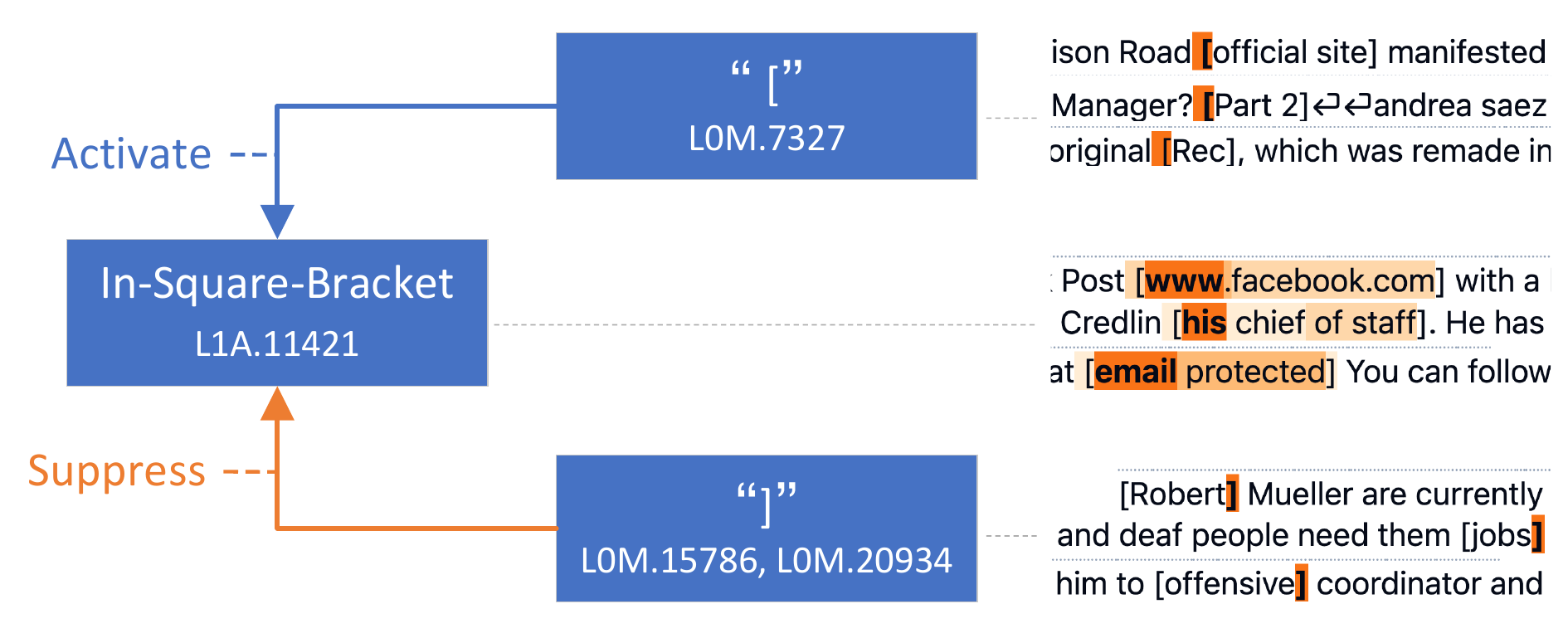}
    }
    \begin{minipage}[b]{\linewidth}
        \centering
        \subfigure[Contribution to a specific \textit{In-Bracket} feature from each token's open or closing bracket features]{
            \label{fig:bracket-L1A11421-contribution}
            \includegraphics[width=0.5\columnwidth]{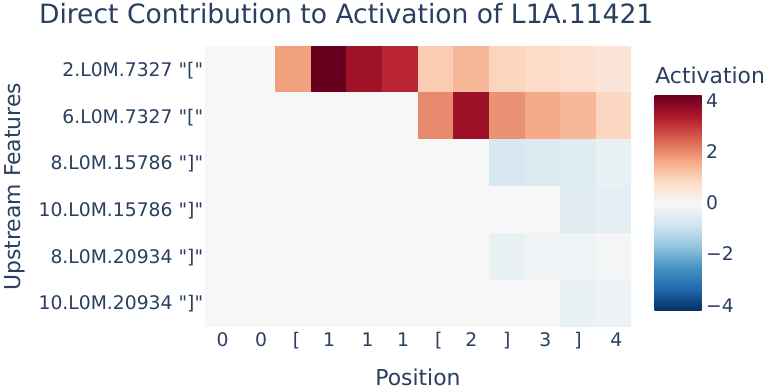}
        }
        \subfigure[Attention Score Trends of a Significant \textit{Bracket Head}]{
            \label{fig:bracket-L1AH1-attn_scores}
            \includegraphics[width=0.3\columnwidth]{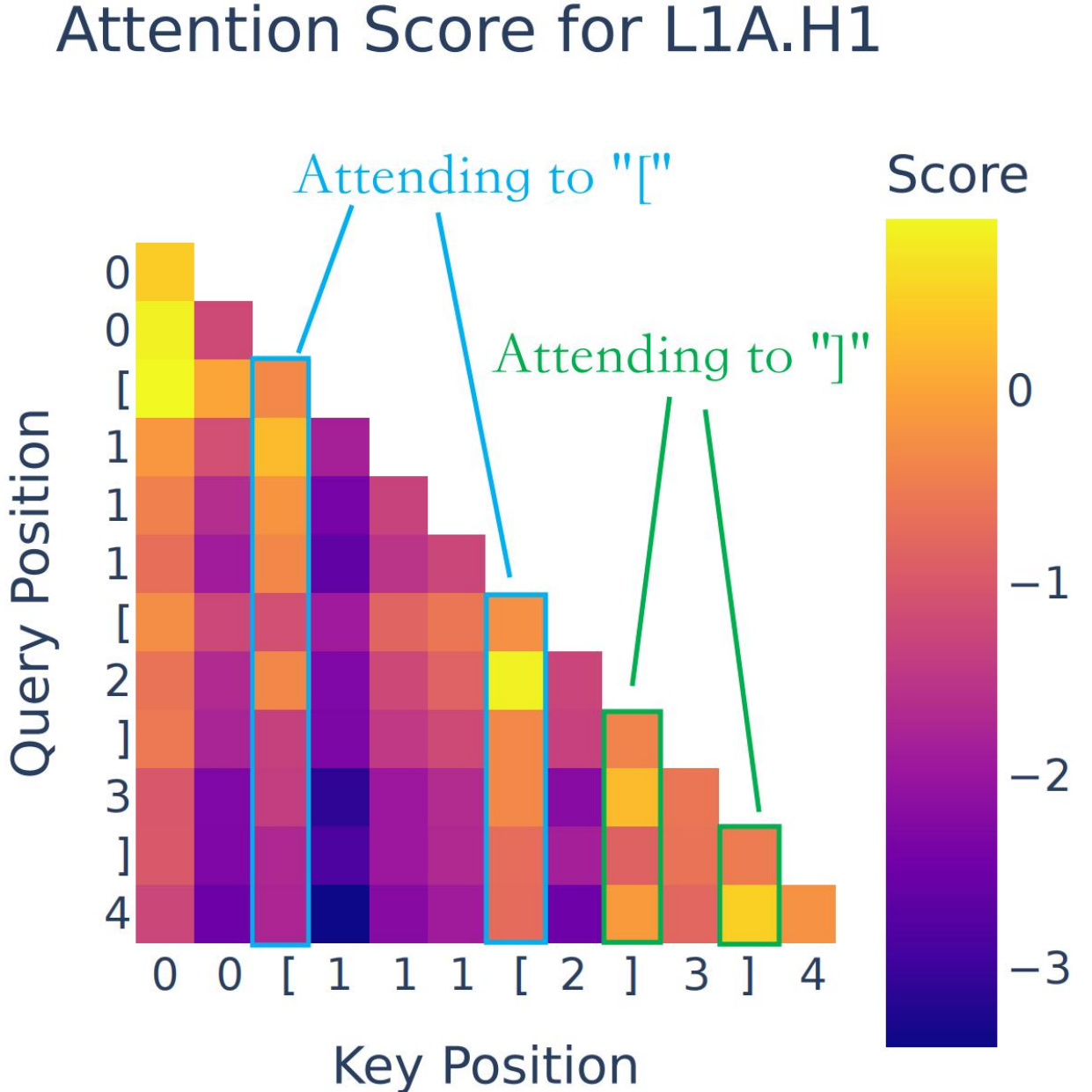}
        }
    \end{minipage}
    
    \caption{(a) \textit{Opening Bracket} features and \textit{Closing Bracket} features have positive and negative contributions to \textit{In-Bracket} features respectively. (b) Closer "\underline{ [}"s activates the \textit{In-Bracket} feature more prominently. (c) Tokens after "\underline{ [}"s start with strong attention to "\underline{ [}"s and become weaker as the sentence continues. This explains the trend in Figure~\ref{fig:bracket-L1A11421-contribution}.}
\end{figure}

We start from a series of \textit{In-Bracket} features in attention blocks of early layers, which activate on tokens inside of brackets, e.g., \textit{deactivated [activated] deactivated}. These features will demonstrate higher activation in deeper nesting of brackets, imitating the behavior of finite state automata~\citep{Bricken2023monosemanticity} with states of bracket nesting hierarchy. We find an \textit{In-Square-Bracket} feature and an \textit{In-Round-Bracket} feature in SAEs trained on layer 1 attention block output, which we call L1A throughout this paper. Since they are at rather early layers, we leverage our Direct Contribution analysis to see how earlier features produce them.

\textbf{Open-bracket features activate in-bracket ones.} Figure~\ref{fig:in-bracket-information-flow} illustrates a simple two-layer bracket circuit in the wild. We inspect contributions to the \textit{In-Square-Bracket} feature in a template, e.g. "0 0 [1 1 1 [2] 3] 4", at token "\underline{1}"s, "\underline{2}", "\underline{~3}" and "\underline{~4}". Experiments show that the activation is mainly promoted by an L0M feature activated by the token "\underline{~[}". It takes on 104.1\%, 102.6\% and 314.2\% of the \textit{In-Square-Bracket} feature's activation respectively at token "\underline{1}", "\underline{2}", and "\underline{~3}", respectively. An average of 83.8\% of these contributions comes through the attention head 1 of L1A, i.e., L1A.H1.

\textbf{Closing-bracket features deactivate in-bracket ones.} The activation of the \textit{In-Square-Bracket} feature is mostly suppressed by a "\underline{]}" feature in L0M (Figure~\ref{fig:bracket-L1A11421-contribution}). The suppression goes through L1A.H1 as well.

\textbf{Interpreting QK attention to "\underline{ [}" and "\underline{]}".} We study the QK circuit of L1A.H1, as shown in Figure~\ref{fig:bracket-L1AH1-attn_scores}. This head attends to "\underline{ [}"s and "\underline{]}"s regardless of the current token. This is mainly caused by $b_Q$ in L1A.H1 attending to the above "\underline{ [}" and "\underline{]}" features.



\subsection{Revisiting Induction Behavior from the SAE Lens}
\label{sec:induction}

\textit{Induction Heads}~\citep{Olsson2022induction} is an important type of compositional circuit with two attention layers which try to repeat any 2-gram that occurred before, i.e. [A][B] ... [A] -> [B]. These circuits are believed to account for most in-context learning functionality in large transformers. Compared to the massive existing literature in understanding the induction mechanism in the granularity of attention heads~\citep{Olsson2022induction, Hendel2023TVInduction, Ren2024SemInduction}, \textit{inter alia}, we seek to present a finer-grained level interpretation of such behavior.

\begin{figure}
    \centering
    \subfigure[Information Flow in Induction Circuit]{
        \label{fig:induction-information-flow}
        \includegraphics[width=0.63\linewidth]{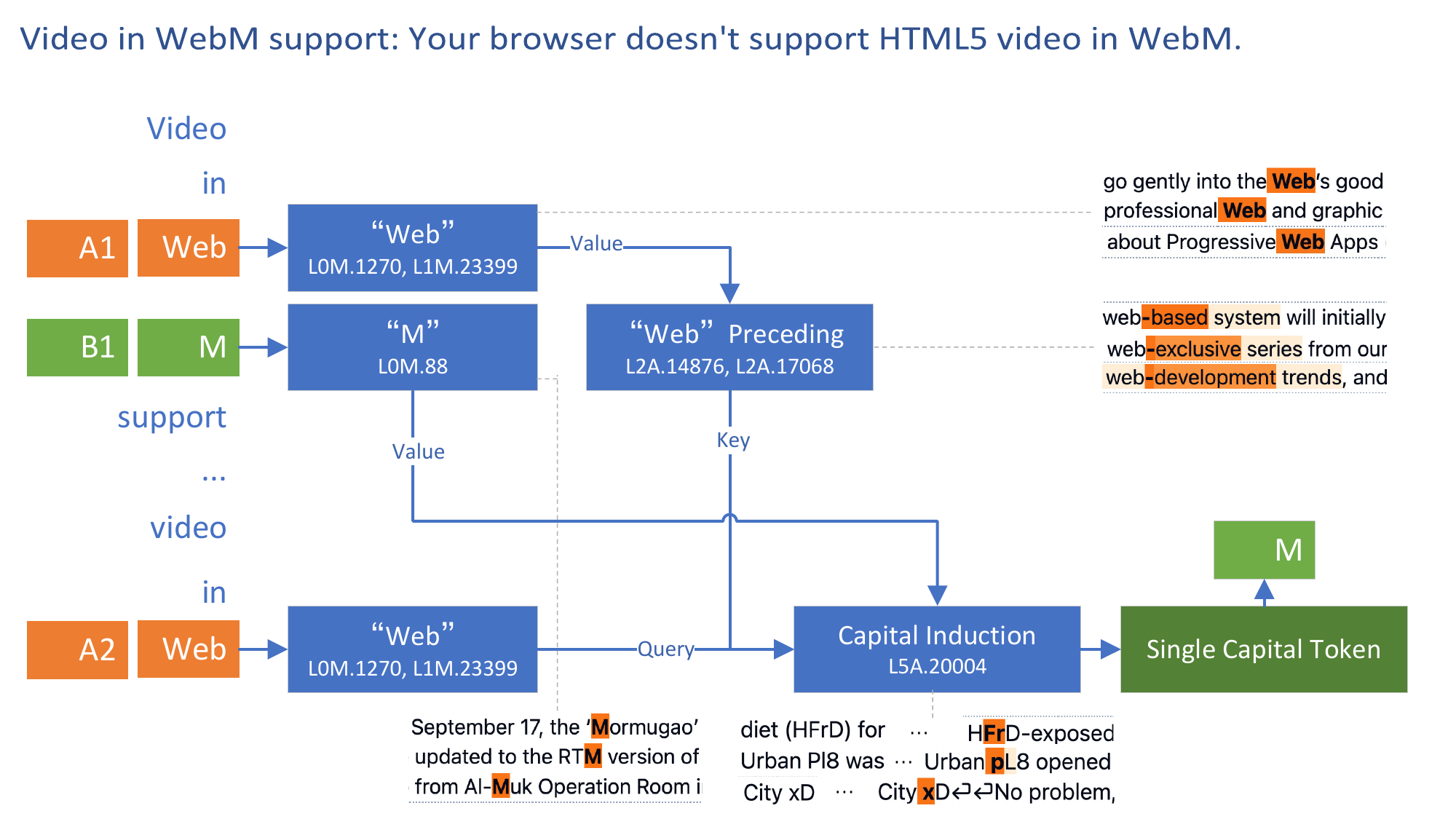}
    }
    \subfigure[QK Top Contributors to a Significant Induction Head]{
        \label{fig:induction-L5AH1Q17K3-attn_scores}
        \includegraphics[width=0.33\linewidth]{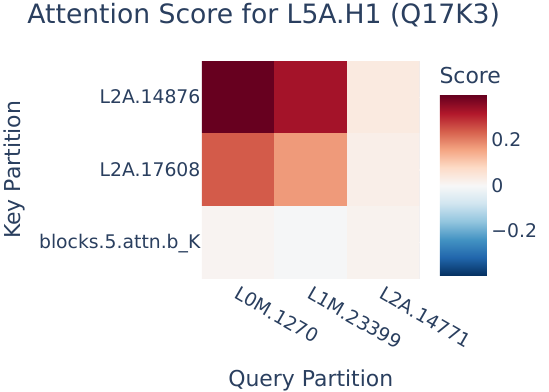}
    }
    
    \caption{"Web"(L0M.1270 and L1M.23399) and "Web" \textit{Preceding} features (L2A.14876 and L2A.17608) jointly lead to QK attention of an induction head. The "M" feature is copied to the last token for the next token prediction.}
\end{figure}

Induction features form a huge feature family. These features are found to be identified by the logit of tokens they enhance through the logit lens~\citep{nostalgebraist2020logitlens}. 
We first study a \textit{Capital Induction} feature contributing to logits of single capital letters on a curated input "Video in WebM support: Your browser doesn't support HTML5 video in WebM." (Figure~\ref{fig:induction-information-flow}). This feature is activated on the second "\underline{ Web}" and amplifies the prediction of "\underline{M}", copying its previous occurrence.

\textbf{Upstream Contribution through OV Circuit} We notice that a series of "\underline{M}" features in the residual stream of the first "\underline{M}" constitute most of the \textit{Capital Induction} feature's activation through OV circuits. L0M.88 takes the lead, which contributes 35.0\% of the feature activation. Auxiliary features from L0A, L1M, and L3M either directly indicate the current token as "\underline{M}" or indicate the current token as a single capital letter. Top 7 of the auxiliary features account for another 33.0\% of the feature activation. Most of these contributions come from L5A.H1, which we along with a concurrent research~\citep{Krzyzanowski2024attention_saes} identify as an induction head.

\textbf{Upstream Contribution to QK Attention} To study how this induction head attends to the first "\underline{M}", we attribute the attention score to upstream feature pairs. The commonality of top contributors is a "\underline{ Web}" feature attending to a "\underline{ Web}" \textit{Preceding} feature (i.e., its previous token is "\underline{ Web}"), as shown in Figure~\ref{fig:induction-L5AH1Q17K3-attn_scores}. 

\textbf{Attributing Preceding features} We further study how "\underline{ Web}" \textit{Preceding} features indicate previous tokens. These contributions mainly come through L2A.H2, which we think to be a previous token head. The relatively high attention score for the previous token can be attributed to a group of L0A features collecting information from positional embeddings.




\section{Revisiting Indirect Object Identification Circuits from the SAE Lens}
\label{sec:ioi_circuits}

\begin{figure}[!ht]
    \centering
    \subfigure[Overview of $s_\text{John}$ circuit]{
        \label{fig:IOI_s_John}
        \includegraphics[width=\linewidth]{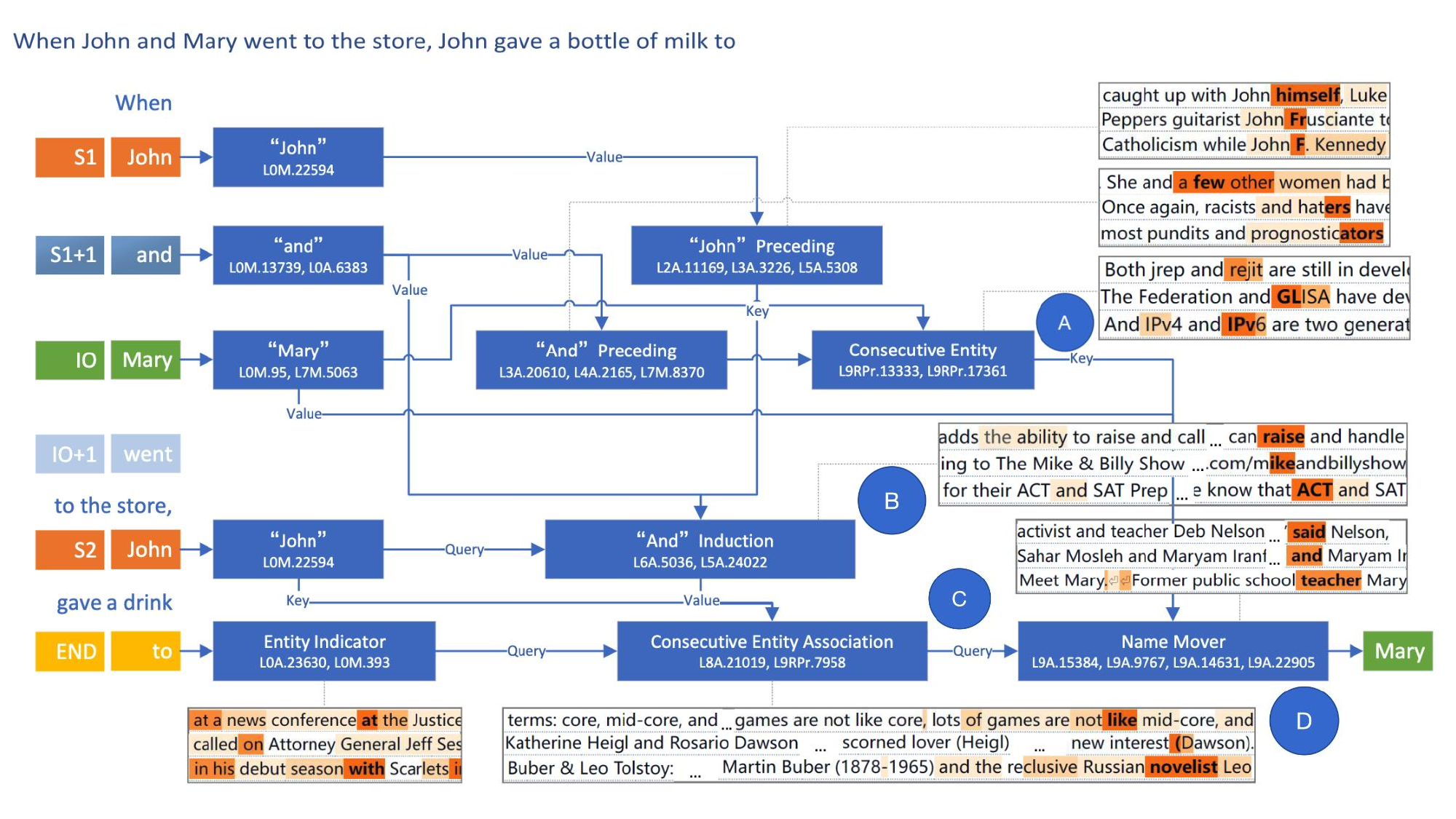}
    }
    \subfigure[A non-rigorous illustration of the key differences between $s_\text{John}$ and $s_\text{Mary}$ circuits]{
        \label{fig:IOI_comparison}
        \includegraphics[width=\linewidth]{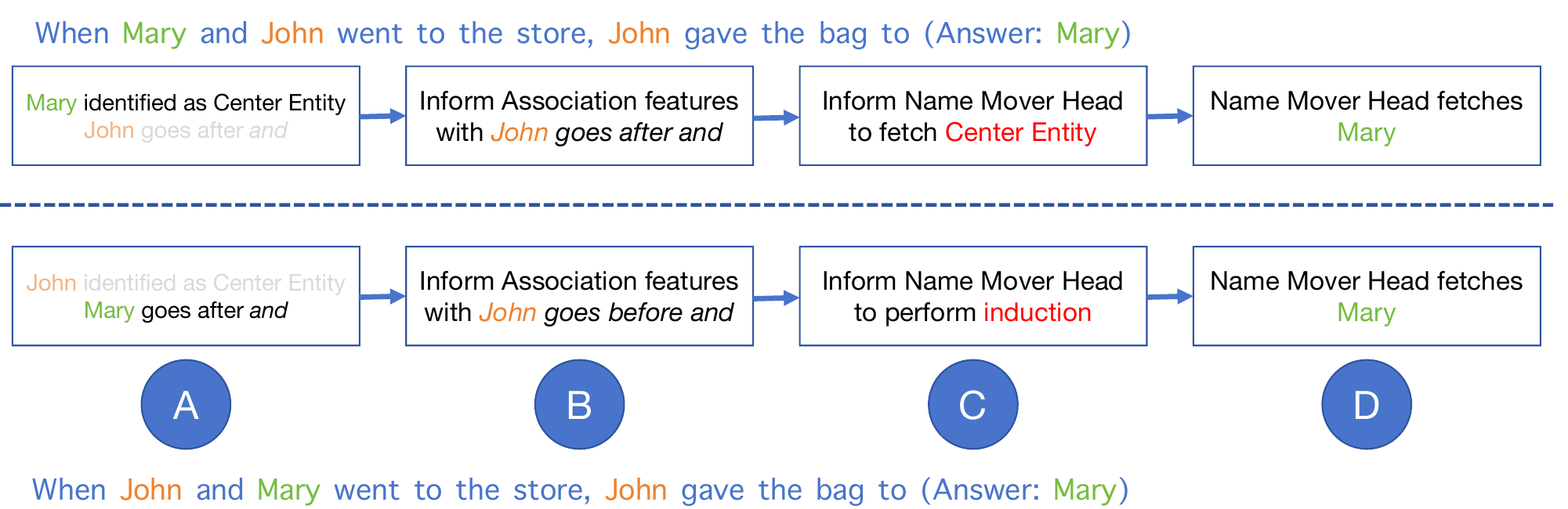}
    }
    
    \caption{In $s_\text{John}$, the consecutive entity feature (denoted as A in Figure~\ref{fig:IOI_s_John}) serves as the key vector for Name Mover Heads to attend to and copy the answer entity to the last token's residual stream. Such a mechanism does not work in $s_\text{Mary}$ because the correct answer is no longer a consecutive entity (i.e., the entity present after the token \textit{and}). See Appendix~\ref{appendix:ioi} for a detailed interpretation of these two examples.}
    \label{fig:IOI_examples}
\end{figure}

For end-to-end circuits in GPT-2 Small, we choose to investigate a task called Indirect Object Identification (IOI)~\citep{Wang2023IOI} with \textit{Hierarchical Attribution}. For instance, GPT-2 can predict "\underline{ Mary}" following the prompt "\underline{When Mary and John went to the store, John gave the bag to}". We call this prompt $s_\text{Mary}$ since it starts with "\underline{ Mary}" and a variant $s_\text{John}$ with a swap in the first two names, i.e. "\underline{When John and Mary went to the store, John gave the bag to}". The answer to both prompts is "\underline{ Mary}", which GPT-2 is able to predict. Existing literature studying this problem does not distinguish between these two templates.
Through the lens of SAE circuits, we validate conclusions in previous work and also discover some subtle mechanistic distinctions in their corresponding circuits.



\subsection{SAE Circuits Closely Agree with Head-Level Ones}

We manage to find the end-to-end information flow in the IOI task example $s_\text{Mary}$ and its variant $s_\text{John}$ with \textit{Hierarchical Attribution}. Then, we identify the pivotal attention heads in the isolated subgraph and attribute their QK scores to earlier SAE features. Discovered SAE feature circuits are of strong consistency with those found based on attention heads: (1) \textit{Name Mover} features correspond to \textit{Name Mover Heads} (L9A.H6, L9A.H9); (2) \textit{Association} features correspond to \textit{S-Inhibition Heads} (L7A.H3, L7A.H6, L8A.H10); (3) \textit{Induction} features correspond to \textit{Induction Heads} (L5A.H5, L6A.H9); (4) \textit{Preceding} features correspond to \textit{Previous Token Heads} (L2A.H2, L3A.H2, L4A.H1).

\subsection{Zooming in on SAE Circuits Yields New Discoveries}

We present a concrete example in the wild that SAE circuits convey more information than their coarse-grained counterparts. We believe this is a positive signal for us to obtain a deeper understanding of language model circuits. Despite the consistency of involved attention heads in $s_\text{John}$ and $s_\text{Mary}$, these two circuits are actually composed of completely different SAE features, as shown in Figure~\ref{fig:IOI_comparison}.

We start with interpreting how GPT-2 predicts " Mary" given the prompt "When John and Mary went to the store, John gave the bag to" ($s_\text{John}$). Though greatly simplified, the information flow is still somehow complicated. We further pick four pivotal feature clusters, as marked in Figure~\ref{fig:IOI_examples}. A non-rigorous interpretation of them is as follows.

\begin{enumerate}[label=\Alph*]
\item " Mary" is recognized as a Consecutive Entity because it occurs after an " and".
\item S2, i.e., the second "John" activates an induction feature. It enhances the logit of "and" though its next token is not.
\item " to" is a representative token indicating the next token is some object or entity. It activates an association feature to retrieve possible entities occurring before. It copies information from feature group B and is informed of the existence of an entity going after an " and".
\item The Name Mover Head receives this information and easily copies the token " Mary" to its residual stream.
\end{enumerate}

The interpretation above highly depends on the fact that the Indirect Object is present after an " and". However, things are quite different in $s_\text{Mary}$ since it comes before the "and". In fact, token " Mary" first activates a Center Entity feature, whose explanation given by GPT-4 is "People or Objects that is likely to be the main topic of the article". The last token still seeks to associate a previously occurring entity but is \textit{informed} to retrieve the Center Entity instead since the Consecutive Entity Association feature has been suppressed by repeated " John"s.

\section{Related Work}
\paragraph{Mechanistic and Representational Interpretability}

Mechanistic Interpretability~\citep{Olah2020zoom, Olah2020early-vision} deems model components, e.g., attention heads and MLP neurons, as \textit{primitives} and explains how they interact with model input and output. This line of research has succeeded in identifying attention-based circuits implementing various NLP tasks~\citep{Olsson2022induction, Wang2023IOI, Heimersheim2023docstring}. Efforts are also made to interpret polysemantic MLP neurons~\citep{Gurnee2023haystack} and editing information stored in MLP parameters~\citep{Meng2022ROMEGPT, Sharma2024ROMEMamba}.

By placing intermediate activations at the center of analysis, Representational Interpretability approaches mostly use linear probes to isolate a targeted behavior in a supervised manner~\citep{Kim2018TCAV, Geiger2023abstraction, Zou2023REPE}. However, such methods may fail to capture unanticipated behaviors.

\paragraph{Sparse Autoencoders} stand in between these two approaches. SAEs disentangle features in the model's \textit{hidden activation}~\citep{Chen2017dict, Subramanian2018dict, Zhang2019dict, Panigrahi2019dict, Yun2021dict, Bricken2023monosemanticity, Cunningham2023SAE} into more interpretable \textit{primitives} than MLP neurons, in an unsupervised manner. Albeit reconstruction errors, \citet{Rajamanoharan2024GatedSAE, Wright2024FeatureSuppression} have proposed to improve SAE training with lower loss and more sparsity.

\paragraph{Circuit Discovery with SAE Features} Previous work mechanistically interprets circuits connecting attention heads and MLP neurons~\citep{Olsson2022induction, Wang2023IOI, Conmy2023ACDC}. As for SAE circuits, \citet{He2024OthelloCircuit} makes a linear approximation of MLP layers by fixing the gate mask of the non-linear activation function; \citet{Marks2024SparseFeatureCircuit} estimates the indirect effect of each SAE feature with attribution patching~\citep{Kramar2024AtP}, which also makes linear assumption of non-linear functions. In contrast, we refactor our computation graph to be completely linear w.r.t. OV and MLP circuits without approximation.

\section{Conclusion and Limitation}
\label{sec:conclusion}

We frame a pipeline to identify fine-grained circuits in Transformer language models. With Sparse Autoencoders and Transcoders, we refactor the model's computation to linear (with respect to a single input). We also propose an efficient approach to isolate subgraphs (i.e. circuits). We showcase that finer-grained circuit analysis provides more beautiful and detailed structures in Transformers. One limitation of our work is that our analysis is specific to certain inputs and might not generalize to other settings. We deem this as a trade-off between granularity and universality. Some extensions can be made to extract more general circuits regarding more abstract behaviors. We leave this for future work.

\newpage

\bibliography{colm2024_conference}
\bibliographystyle{colm2024_conference}

\newpage

\appendix

\section{Sparse Autoencoder Training}
\label{appendix:sae_training}

We trained an SAE (Section~\ref{sec:sparse_autoencoder}) on each of the outputs of the 12 attention layers and 24 residual stream activation (before entering attention layers and MLP layers). We trained a Skip SAE (Section~\ref{sec:skip_sae}) through each MLP layer, using residual stream activation before MLP as input and MLP output activation as the label. Here are our training settings:

\begin{itemize}
    \item Each SAE has 24,576 dictionary features, which is 32 times the hidden dimension of GPT-2 Small.
    
    \item We use the Adam optimizer with a learning rate of 4e-4 and betas of (0, 0.9999) for 1 billion tokens from the OpenWebText corpus. We trained against a reconstruction loss (measured by MSE of input and reconstructed output), a sparsity loss (proxied by the L1 norm of the feature activations, with a coefficient of 8e-5 (1.2e-4 for attention output SAEs)), and a ghost gradient loss. A batch size of 4,096 is used. We use an NVIDIA A100-80GB GPU for training of each SAE, which lasts for 20 hours.
    
    \item The first 256 tokens of each sequence are used as input, discarding the remaining tokens and sequences shorter than 256 tokens. Generated activations are shuffled actively in an activation buffer.
    
    \item We normalize the input activations to have a norm of the square root of LM hidden size (i.e., $\sqrt{768}$ for GPT-2 Small). We further normalize the MSE loss by the variance of output along the hidden dimension (a bit like the latter part in LayerNorm, except that we're not taking the mean of output).

    $$
        \mathcal{L}_\text{MSE}=(x_\text{normed}-\hat{x}_\text{normed}) /
          \lVert \hat{x}_\text{normed} - \bar{\hat{x}}_\text{normed} \rVert_2
    $$

    \item We use untied weights for the encoder and decoder. Decoder bias (or pre-encoder bias) is removed (for the sake of simpler circuit analysis). Decoder norms are reset to less than or equal to 1 after each training step.

    \item \textbf{*}We prune the dictionary features with a norm less than 0.99, max activation less than 1, and activation frequency less than 1e-6 after training.

    \item \textbf{*}We finetuned the decoder and a feature activation scaler of the pruned SAEs on the same dataset to deal with feature suppression.
\end{itemize}

\subsection{Feature Pruning}

Some of the SAE features obtained from end-to-end training are too sparse (i.e., can hardly be activated) to reflect a certain aspect of the input corpus. These features are more like "local codes" (in neuroscience). They are activated by very specific tokens. These features are trivial and not helpful for understanding an activation pattern from a compositional perspective. Feature pruning aims to remove these trivial features and keep the more meaningful ones.

In practice, a dictionary feature will be pruned if it meets one of the following criteria:

\paragraph{Norm less than 0.99:}

In SAE training, we use an L1 loss as a differentiable approximation of L0 loss, to encourage sparsity in the feature activations. The side effect is that the L1 loss as well encourages a lower value of the feature activations and a larger feature norm. Thus, if a feature is really "useful" in reconstructing the input, it should have a norm as large as possible. We prune the features without the tendency to grow.

\paragraph{Max activation less than 1:}

Given a fixed norm of the feature, a feature with a low max activation value contributes little to reconstructing the input. We find this kind of feature activated in some non-related situations and thus non-interpretable. We empirically set the threshold to 1 and prune the features below it.

\paragraph{Activation frequency less than 1e-6:}

A feature with an ultra-low activation frequency is considered too local to be useful. We find that these features often correspond to some specific tokens in some specific contexts, which is too trivial to be recognized as a feature. We empirically set the threshold to 1e-6 and prune the features activated at a frequency below it.

\subsection{Finetuning against Feature Suppression}

Feature suppression refers to a phenomenon where loss function in SAEs pushes for smaller feature activation values, leading to suppressed features and worse reconstruction quality. Wright and Sharkey deduced that for an L1 coefficient of $c$ and dimension $d$, instead of having a ground truth feature activation of $g$, the optimal activation SAEs may learn is $g - \frac{cd}{2}$.

To address this issue, we finetune the decoder and a feature activation scaler of the pruned SAEs on the same dataset. Only the reconstruction loss (i.e., the MSE loss) is applied in this fine-tuning process. Encoder weights are fixed during this process to keep the sparsity of the dictionary. Finetuning may also repair flaws introduced in the pruning process and improve the overall reconstruction quality.

\subsection{Statistics of Sparse Autoencoders}

We evaluate the L0 loss, variance explained, and reconstruction CE loss of each trained SAE. The L0 loss computes the average feature activated at each token. Variance explained computes

$$
    EV = 1 - \frac{\lVert \hat{y}-y \rVert_2^2}{\sigma^2(y)},
$$

which measures the proportion to which an SAE accounts for the activation variation. Reconstruction CE loss is the final cross-entropy loss of the language model, where the activation is replaced with the SAE reconstructed one. The reconstruction CE score shows how good the reconstruction CE loss is w.r.t the original CE loss and the ablated CE loss by computing

$$
    s = \frac{\mathcal{L}_\text{recons}-\mathcal{L}_\text{ablate}}{\mathcal{L}_\text{original}-\mathcal{L}_\text{ablate}},
$$

where $\mathcal{L}_\text{recons}$, $\mathcal{L}_\text{original}$ and $\mathcal{L}_\text{ablate}$ refer to the reconstruction CE loss, the original CE loss and the ablated CE loss respectively.

The statistics of each SAE is as shown in Table.~\ref{table:stat_attn_sae}, Table.~\ref{table:stat_mlp_sae} and Table.~\ref{table:stat_resid_sae}.

\begin{table}
    \caption{Statistics of Attention Output SAEs}
    \label{table:stat_attn_sae}
    \begin{tabular}{lllll}
        \toprule
        SAE & Var. Explained & L0 Loss & Reconstruction CE Score & Reconstruction CE Loss \\
        \midrule
        L0A & 92.25\% & 29.66 & 99.24\% & 3.2327 \\
        L1A & 82.48\% & 65.57 & 97.19\% & 3.2138 \\
        L2A & 83.39\% & 69.85 & 94.29\% & 3.2150 \\
        L3A & 69.23\% & 53.59 & 87.00\% & 3.2173 \\
        L4A & 74.91\% & 87.35 & 89.99\% & 3.2171 \\
        L5A & 82.12\% & 127.18 & 97.81\% & 3.2145 \\
        L6A & 76.63\% & 100.89 & 94.31\% & 3.2158 \\
        L7A & 78.51\% & 103.30 & 91.32\% & 3.2182 \\
        L8A & 79.94\% & 122.46 & 88.67\% & 3.2172 \\
        L9A & 81.62\% & 107.81 & 89.55\% & 3.2187 \\
        L10A & 83.75\% & 100.44 & 87.70\% & 3.2201 \\
        L11A & 84.81\% & 22.69 & 85.49\% & 3.2418 \\
        \bottomrule
    \end{tabular}
\end{table}

\begin{table}
    \caption{Statistics of MLP Transcoders}
    \label{table:stat_mlp_sae}
    \begin{tabular}{lllll}
        \toprule
        SAE & Var. Explained & L0 Loss & Reconstruction CE Score & Reconstruction CE Loss \\
        \midrule
        L0M & 94.16\% & 19.59 & 99.65\% & 3.1924 \\
        L1M & 82.02\% & 48.63 & 86.35\% & 3.1816 \\
        L2M & 86.32\% & 50.90 & 81.24\% & 3.1851 \\
        L3M & 76.55\% & 56.91 & 83.43\% & 3.1867 \\
        L4M & 73.38\% & 76.03 & 80.08\% & 3.1888 \\
        L5M & 73.49\% & 84.11 & 84.18\% & 3.1881 \\
        L6M & 72.79\% & 90.34 & 82.85\% & 3.1912 \\
        L7M & 73.18\% & 86.38 & 81.89\% & 3.1911 \\
        L8M & 74.14\% & 87.29 & 83.25\% & 3.1913 \\
        L9M & 75.89\% & 90.08 & 81.89\% & 3.1930 \\
        L10M & 79.66\% & 94.85 & 81.60\% & 3.1987 \\
        L11M & 80.33\% & 79.12 & 77.33\% & 3.2169 \\        
        \bottomrule
    \end{tabular}
\end{table}

\begin{table}
    \caption{Statistics of Residual Stream SAEs}
    \label{table:stat_resid_sae}
    \begin{tabular}{lllll}
        \toprule
        SAE & Var. Explained & L0 Loss & Reconstruction CE Score & Reconstruction CE Loss \\
        \midrule
        L0RPr & 98.98\% & 6.89 & 99.90\% & 3.1907 \\
        L0RM & 95.94\% & 42.50 & 99.34\% & 3.2658 \\
        L1RPr & 96.98\% & 21.96 & 99.62\% & 3.1935 \\
        L1RM & 95.53\% & 34.11 & 99.77\% & 3.2133 \\
        L2RPr & 96.03\% & 28.18 & 99.01\% & 3.2268 \\
        L2RM & 94.45\% & 40.17 & 99.32\% & 3.2662 \\
        L3RPr & 94.43\% & 38.22 & 98.95\% & 3.2867 \\
        L3RM & 93.13\% & 48.44 & 99.13\% & 3.2673 \\
        L4RPr & 92.08\% & 49.19 & 99.31\% & 3.2782 \\
        L4RM & 91.00\% & 61.66 & 99.26\% & 3.2771 \\
        L5RPr & 90.68\% & 60.34 & 99.09\% & 3.2950 \\
        L5RM & 89.90\% & 76.22 & 99.11\% & 3.2839 \\
        L6RPr & 90.03\% & 70.06 & 98.93\% & 3.2899 \\
        L6RM & 89.57\% & 88.95 & 98.59\% & 3.2830 \\
        L7RPr & 88.86\% & 79.91 & 98.88\% & 3.2943 \\
        L7RM & 88.28\% & 98.60 & 98.94\% & 3.2828 \\
        L8RPr & 87.99\% & 89.37 & 98.55\% & 3.2952 \\
        L8RM & 87.32\% & 108.72 & 98.70\% & 3.2863 \\
        L9RPr & 87.38\% & 100.68 & 99.17\% & 3.2938 \\
        L9RM & 86.66\% & 119.59 & 98.15\% & 3.2889 \\
        L10RPr & 86.72\% & 115.35 & 98.59\% & 3.2984 \\
        L10RM & 86.07\% & 126.19 & 98.14\% & 3.3036 \\
        L11RPr & 85.76\% & 120.86 & 97.93\% & 3.3212 \\
        L11RM & 85.40\% & 94.20 & 98.42\% & 3.3910 \\        
        \bottomrule
    \end{tabular}
\end{table}

\section{General Direct Contribution Computation}

In Sec.~\ref{sec:attn_contribution} and Sec.~\ref{sec:skip_sae}, we have shown how we compute direct contribution towards attention outputs, attention scores, and SAE feature activation, which is a linear effect of each input partition. However, it may still remain confusing why we can compute a linear contribution in such non-linear functions as attention blocks. For a clarification of how direct contribution works, we introduce our general mathematical formation of direct contribution computation in this section.

The term \textbf{direct contribution} refers to how partitions of upstream model activations respectively contribute to the downstream (through only direct ways, e.g. a single model layer), and constitute the downstream model activations. We start from linear functions, which are the simplest case of direct contribution computation. Given a model activation $x\in\mathbb{R}^H$ and its arbitrary n-parted partition $x=\sum_{i=1}^n v_i$, where $v_i\in\mathbb{R}^H$ is the $i$-th partition of $x$. For any affine transformation $f:\mathbb{R}^H\to\mathbb{R}^K$ mapping $x$ to a downstream activation $f(x)=Wx+b$, $W\in\mathbb{R}^{K \times H}$, $b\in\mathbb{R}^K$, we have

\begin{equation}
    f(x)=W\sum_{i=1}^n v_i + b=\sum_{i=1}^n Wv_i+b,
\end{equation}

from which we learned that each partition $v_i$ separately contributes to $f(x)$ by $Wv_i$ (since it's the only term related to $v_i$ in the final summation, and the bias $b$ contributes to $f(x)$ by its own value $b$. This contribution ribution is natural thanks to the additive (w.r.t vector addition) nature of linear mapping.

\begin{figure}
    \centering
    \includegraphics[width=0.9\linewidth]{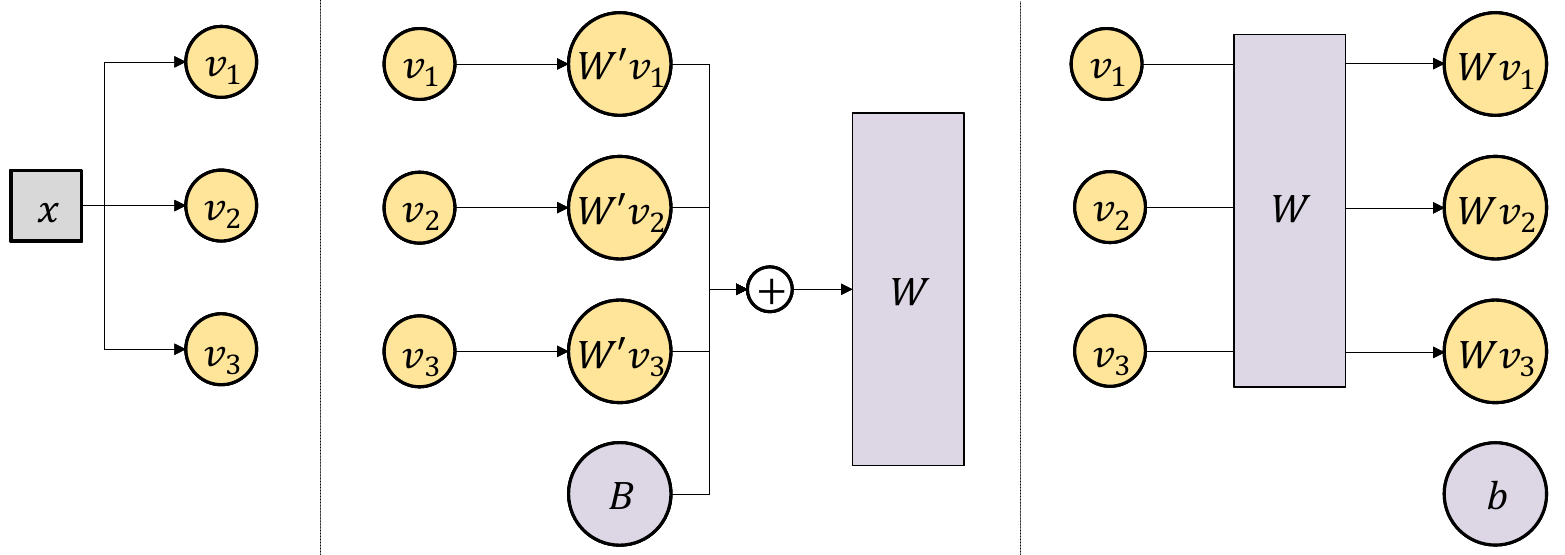}
    \caption{The workflow of interpreting a non-linear transformation where the transformation matrix can be linearly decomposed. We first compute the direct contribution $W'v_i$ to the transformation matrix $W$ of each partition $v_i$ of $x$ to reveal the formation of $W$, and then treat the computed $W$ as constant to compute the final direct contribution $Wv_i$.}
    \label{fig:math_framework}
\end{figure}

Nevertheless, computation in practical neural networks is often much complicated than the above affine transformation or its simple nesting. Non-linear transformation (e.g. \verb|LayerNorm|, \verb|Softmax|, \verb|ReLU|) is ubiquitous. We cannot simply ignore these non-linear operators since the powerful fitting capacity of neural networks often just comes from the non-linear parts. To deal with these non-linear transformations, we propose a more general direct contribution computing strategy. For any transformation $f:\mathbb{R}^H\to\mathbb{R}^K$ where $f$ has a form of $f(x)=W(x)x+b$, $W:\mathbb{R}^H\to\mathbb{R}^{K \times H}$, $b\in\mathbb{R}^K$, we have

\begin{equation}
    \label{eq:nonlinear_contribution}
    \begin{aligned}
        f(x)=W(x)\sum_{i=1}^n v_i+b=\sum_{i=1}^n W(x)v_i+b,
    \end{aligned}
\end{equation}

where we treat $W(x)$ as a constant linear transformation matrix. Then, we can claim that $i$-th partition $v_i$ contributes to the result $f(x)$ by $W(x)v_i$ through the posterior linear transformation with a constant $W(x)$. We must state this contribution computation is nothing but trivial if we don't further interpret how partitions affect $W(x)$ and the related impact to the following transformation or further restrict the $W(x)$to make sure it's just close to a constant or its variation is unimportant. Thus, for $W$ that having a similar form as $f$, e.g. $W(x)=W'(x)x+B$, $W':\mathbb{R}^H\to\mathbb{R}^{(K \times H) \times H}$, $B\in\mathbb{R}^{K\times{}H}$, we can iteratively apply the linear decomposition Eq.~\ref{eq:nonlinear_contribution} to $W$ (which we use to interpret attention pattern in Sec.~\ref{sec:attn_contribution}),

\begin{equation}
    \begin{aligned}
        W(x)=W'(x)\sum_{i=1}^n v_i+B=\sum_{i=1}^n W'(x)v_i+B
    \end{aligned}
\end{equation}

The above transformations could be nested to compute direct contribution to further activations. Take $f=f_1 \circ f_2$ as a twofold nesting example, where $f_1(x)=W_1x+b_1$ and $f_2(x)=W_2(x)x+b_2$, it can be easily induced that

\begin{equation}
    \label{eq:nonlinear_contribution_matrix}
    \begin{aligned}
        f(x)=W_1W_2(x)\sum_{i=1}^n v_i+W_1b_2+b_1=\sum_{i=1}^n W_1W_2(x)v_i+W_1b_2+b_1,
    \end{aligned}
\end{equation}

and get the respective contribution of every $v_i$ and $b_i$. Direct contribution through deeper nested transformations can be computed in similar ways.

As a brief summary, the core idea of direct contribution computation for any non-linear function is to first compute how the non-linear part is formed w.r.t each input partition by iteratively applying direct contribution computation, and then consider the non-linear part as determined, regard the function to be linear, and compute a linear contribution to the function output. We usually allow the determined non-linear part to go through a simple extra activation function like \verb|Softmax| or \verb|ReLU|, since this will not undermine the understanding of this non-linear part. This workflow can be applied to non-linear functions like bi-linear functions and attention.







\section{Hierachical Attribution Algorithm}
\label{appendix:hierachical_attribution}

In this section, we introduce the detailed implementation of the Hierarchical Attribution algorithm to obtain a subgraph $G'$ from the original computational graph $G$ with threshold $\tau$, as shown in Algorithm~\ref{alg:hierachical_attribution}.

\begin{algorithm}
\caption{Hierachical Attribution}
\label{alg:hierachical_attribution}
\begin{algorithmic}
\Require $\tau>0,G,t$ \Comment{$t$ for the root node}
\Ensure Optimized subgraph $G'$

\State $N' \gets \varnothing$
\ForAll{$v$ in reversed topological sort of $G$}
    \If{$v=t$}
        \State $v.grad \gets 1$
    \Else
        \State $v.grad \gets 0$
        \ForAll{$u$ in direct successors of $v$ in $G$}
            \State $v.grad \gets v.grad + \nabla_{a_v}a_u \cdot u.grad$ \Comment{Do normal back-propagation}
        \EndFor
        
        \If{$v.grad\cdot a_v < \tau$}
            \State $v.grad\gets 0$
            \State $\operatorname{attr}_v\gets 0$
        \Else
            \State $\operatorname{attr}_v\gets v.grad\cdot a_v$
            \State $N' \gets N' \cup \{v\}$
        \EndIf
    \EndIf
\EndFor
\State $G'\to G[N']$
\end{algorithmic}
\end{algorithm}

Afterwards, we can compute $ G'$'s contribution by adding up the attribution scores of all its leaf nodes.

\section{Equality of Output Activation and Leaf Nodes Attribution}
\label{appendix:attribution_equality}

We demonstrate the proof for Theorem~\ref{theorem:attribution_equality} as below, which is quite simple:

\begin{proof}
For any activated node $u$ (i.e., $a_u > 0$), it holds that

\begin{equation}
    \label{eq:decompose_activated_node}
    \begin{aligned}
        a_u&=\operatorname{ReLU}\left(\sum_{v\to u\in E} k_{v,u} a_v\right)\\
        &=\sum_{v\to u\in E} k_{v,u} a_v\\
        &=\sum_{v\to u\in E, a_t>0} k_{v,u} a_v
    \end{aligned}
\end{equation}

By iteratively applying Eq.~\ref{eq:decompose_activated_node}, we can obtain

\begin{equation}
    \begin{aligned}
        a_t&=\sum_{\deg_{\text{in}}(v)=0,a_v>0} a_v \cdot\nabla_{a_t}a_v\\
        &=\sum_{\deg_{\text{in}}(v)=0} a_v \cdot\nabla_{a_t}a_v\\
        &=\sum_{\deg_{\text{in}}(v)=0} \operatorname{attr}_{v,t}
    \end{aligned}
\end{equation}
\end{proof}

\section{Additional Explanation of IOI Circuit}
\label{appendix:ioi}

We further explain the feature circuit we discovered in $s_\text{Mary}$ and $s_\text{John}$, by listing the meaning or functionality of pivotal features in these two exemplars.

\begin{figure}
    \centering
    \includegraphics[width=\linewidth]{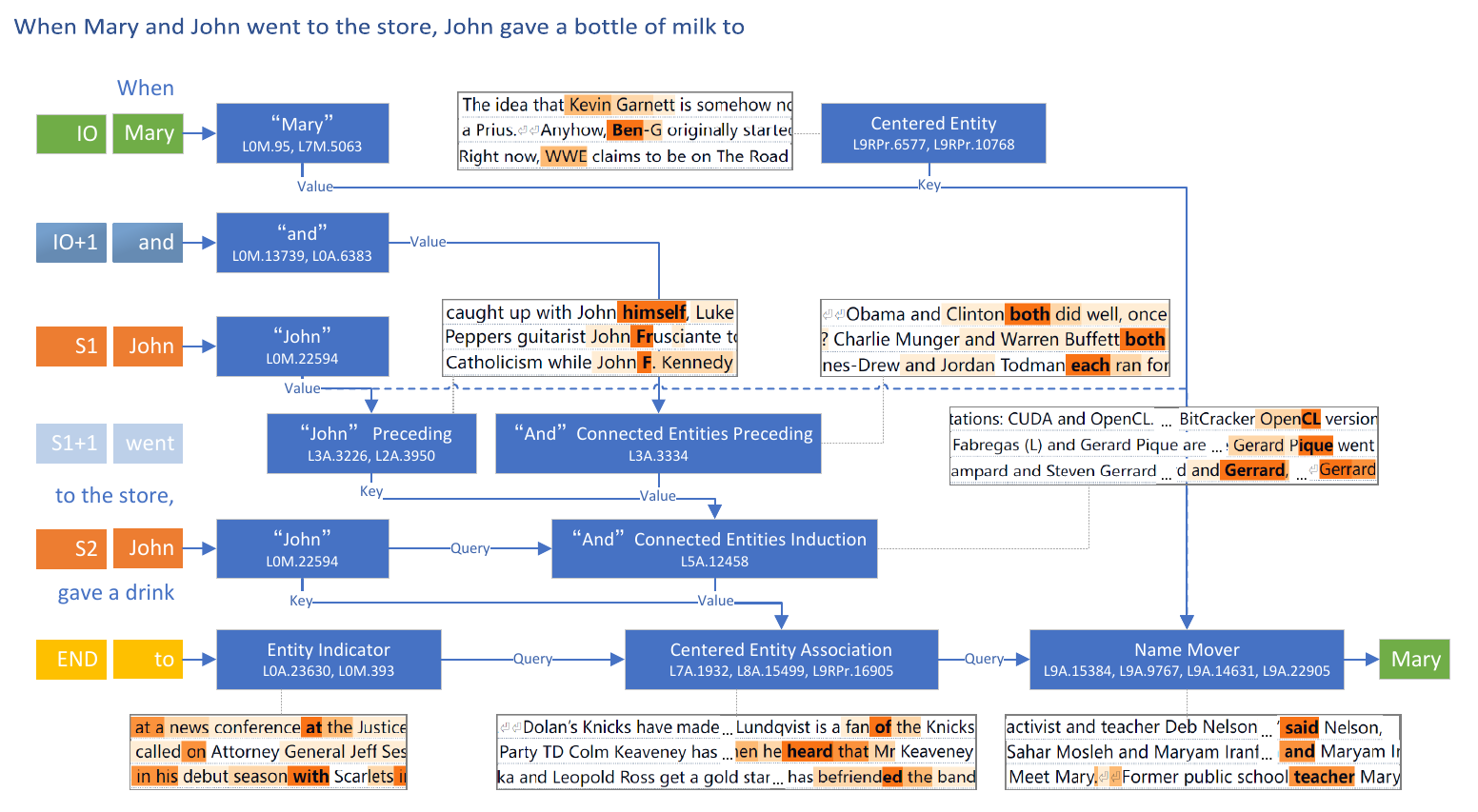}
    
    \caption{Overview of $s_\text{Mary}$ circuit.}
    \label{fig:IOI_s_Mary}
\end{figure}

The pivotal features in $s_\text{John}$ (Figure~\ref{fig:IOI_s_John}):

\begin{itemize}
    \item "John", "and" and "Mary" features simply imply the current token as "John", "and", and "Mary";
    \item \textit{Entity Indicator} features are activated on prepositions or transitive verbs, indicating that its next token will likely be an entity.
    \item "John" \textit{Preceding} features collect information from the previous token and imply its previous token as "John";
    \item "And" \textit{Preceding} features collect information from the previous token and imply its previous token as "and";
    \item \textit{Consecutive Entity} features are a mixture of "Mary" features and "And" \textit{Preceding} features imply the current token as the [B] part of an [A] and [B] pattern, where [A] and [B] serve as entities.
    \item "And" \textit{Induction} features attend to "and" (by matching \textit{S1} and \textit{S2}), and collects the "and" information from \textit{S1+1}, implying there's an "and" goes after "John".
    \item \textit{Consecutive Entity Association} features take advantage of the structural information from "And" \textit{Induction} features, and decide to retrieve the entity lying after "and", by attending to \textit{Consecutive Entity} features in \textit{Name Mover Heads}.
    \item \textit{Nave Mover} features conduct the final step to move the "Mary" information from the targeted \textit{Consecutive Entity} token.
\end{itemize}

The pivotal features in $s_\text{Mary}$ (Figure~\ref{fig:IOI_s_Mary}):

\begin{itemize}
    \item "John", "and", "Mary", \textit{Entity Indicator} and "John" \textit{Preceding} features play the same role as in $s_\text{John}$.
    \item \textit{Centered Entity} features are activated at the first occurrence of a seemingly important name or object, marking it out for potential future reference.
    \item "And"-\textit{Connected Entities Preceding} features collect information from several previous tokens (mainly the token "and") and imply there's an [A] and [B] pattern before this token.
    \item "And"-\textit{Connected Entities Induction} features collect information from "And"-\textit{Connected Entities Preceding}, again by matching \textit{S1} and \textit{S2}.
    \item \textit{Centered Entity Association} features take advantage of the structural information from "And"-\textit{Connected Entities Induction} features and decide to retrieve the entity lying before "and", by attending to \textit{Centered Entity} features in \textit{Name Mover Heads}. This behavior is not completely symmetrical to that with \textit{Consecutive Entity} features since \textit{Centered Entity} features do not know about the token "and" after it. However, this behavior is still reasonable since if there's another \textit{Centered Entity} before \textit{IO}, then this entity can be another correct answer.
    \item \textit{Nave Mover} features again conduct the final step to move the "Mary" information from the targeted \textit{Consecutive Entity} token.
\end{itemize}

\end{document}